\documentclass[runningheads]{llncs}

\usepackage[T1]{fontenc}
\usepackage{graphicx}
\usepackage{booktabs}
\usepackage{amsmath,amssymb}
\usepackage{multirow}
\usepackage{xcolor}

\begin{document}

\title{In Defense of OCTA: The Reconstruction-Utility Gap in OCT-to-OCTA Synthesis}
\titlerunning{In Defense of OCTA}

\author{Michael Chertok\inst{1} \and Alon Tiosano\inst{2} \and Orly Gal-Or\inst{2} \and
Lior Kramarski\inst{2} \and Einav Baharav Shlezinger\inst{2} \and Irit Bahar\inst{2} \and
Lior Wolf\inst{3}}
\authorrunning{M. Chertok et al.}
\institute{MarkerLeap, Raanana, Israel \\ \email{michael@markerleap.com}
\and Rabin Medical Center, Petah Tikva, Israel
\and Tel Aviv University, Tel Aviv, Israel}

\maketitle

\begin{abstract}
Optical coherence tomography angiography (OCTA) images retinal blood flow, giving capillary-perfusion and
foveal-avascular-zone biomarkers that grade diabetic-retinopathy ischemia. Because OCTA hardware is less
common than structural OCT, recent work synthesizes it from OCT, reporting strong reconstruction (3D PSNR
$>31$~dB, SSIM $>0.9$). We ask not whether the synthetic image looks similar, but whether it supports the
measurements OCTA is acquired for. A frozen \emph{real}-OCTA segmenter, applied as a probe to two synthesizers
(XOCT, TransPro), shows downstream Dice falling with structural fineness: large vessels survive ($0.862 \to
0.831$) while the fine capillary network collapses ($0.798 \to 0.635$, five times the large-vessel loss; paired
Wilcoxon $p<10^{-3}$), TransPro worse throughout. A matched-blur control shows this detail is fabricated, not
blurred. Retrained on a private Spectralis dataset, neither synthesizer reproduces the neovascular lesion
(qualitative, $n{=}3$). Reconstruction fidelity is not clinical utility; we establish downstream-task fidelity as
the evaluation OCT-to-OCTA synthesis needs.
\end{abstract}

\section{Introduction}
\label{sec:intro}

OCTA reconstructs retinal blood flow from motion contrast between repeated OCT B-scans, producing
depth-resolved maps of the superficial, intermediate, and deep vascular plexuses (the layered capillary
networks of the retina), the foveal avascular zone (FAZ, the normally capillary-free region at the foveal
center), and the normally avascular outer retina where neovascular lesions appear~\cite{spaide2018octa}. These
maps supply vascular biomarkers used in retinal disease assessment: capillary dropout (loss of perfused
capillaries) and FAZ enlargement indicate macular ischemia (reduced macular blood supply) in diabetic retinopathy,
while macular neovascularization (MNV), abnormal new vessels growing
under or into the macula, is the defining lesion of neovascular age-related macular degeneration (nAMD), and
its exudative activity (intraretinal or subretinal fluid) triggers treatment. OCTA hardware is far less widespread than
structural OCT, which has motivated a line of work that \emph{synthesizes} OCTA from OCT
alone~\cite{li2024transpro,khosravi2025xoct,chen2025mutri,badhon2026bbdm}. These methods report increasing
reconstruction fidelity, peak signal-to-noise ratio (PSNR) above $31$~dB in 3D and structural similarity
(SSIM) above $0.9$, and are naturally read as evidence that OCTA acquisition could be replaced by synthesis
from a cheaper OCT-only device.

We show this reading is not yet warranted. Reconstruction metrics measure pixel agreement, not whether the
synthetic image supports the downstream vascular measurements OCTA exists to provide. A synthesizer can score
well on PSNR by reproducing the dominant large vessels and smooth background while fabricating or omitting the
fine capillary detail that carries the diagnosis, and PSNR, dominated by low spatial frequencies, will not
reveal it. The clinical question is not ``how similar is the synthetic image'' but ``can you make the same
measurement from it.''

\begin{figure}[!t]
\centering
\includegraphics[width=0.84\textwidth]{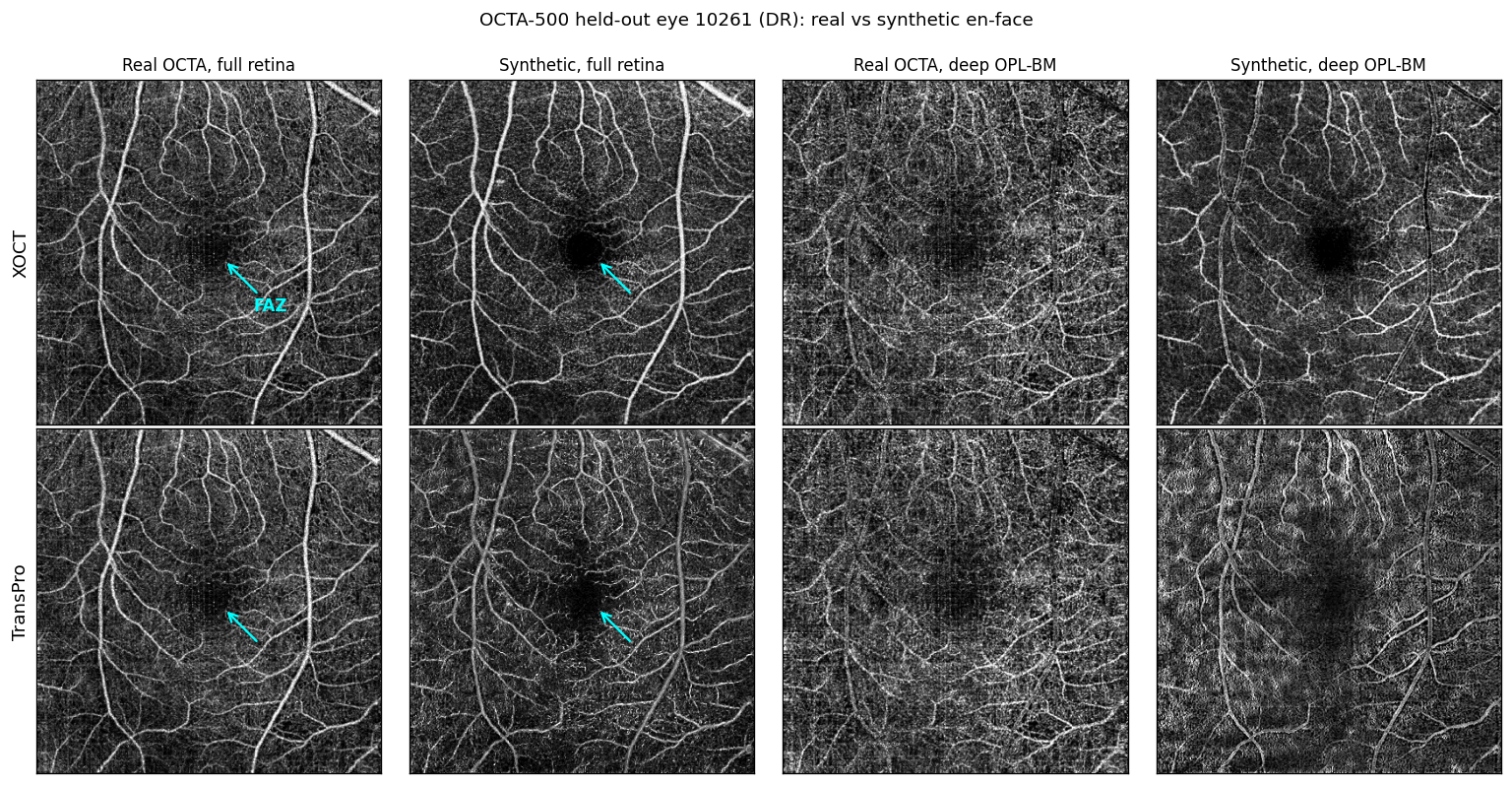}
\caption{Two synthesizers erode the diagnostic microvasculature at high pixel fidelity (OCTA-500 held-out eye,
diabetic retinopathy). Rows: XOCT and TransPro. Columns: real full-retina en-face, synthetic full-retina, real
deep OPL--BM slab (outer plexiform layer to Bruch's membrane), synthetic deep OPL--BM. Both reproduce the large
arcade vessels, which drives high PSNR/SSIM, but the synthetic full-retina enlarges and over-regularizes the
foveal avascular zone (FAZ, the capillary-free foveal center read for ischemia; cyan marker) into a round dark
void and degrades the fine perifoveal capillaries. In the deep OPL--BM slab, both real and synthetic are
dominated by projection tails (decorrelation shadows cast by superficial vessels~\cite{spaide2015artifacts});
the synthesizer reproduces this projected skeleton but not distinct deep-layer detail. The metrics reward the
large vessels the models get right and are blind to the capillaries and FAZ they erase.}
\label{fig:qual}
\end{figure}

We answer that question with a fixed probe: a frozen segmentation network trained on \emph{real} OCTA, fed
synthetic OCTA from two state-of-the-art methods, XOCT and TransPro. The drop in segmentation quality,
stratified by vascular structure, is the utility gap. On OCTA-500 (Fig.~\ref{fig:qual}, Table~\ref{tab:strat})
both synthesizers collapse the finer structure at matched reconstruction fidelity, capillary Dice falls worst;
XOCT preserves large vessels (five times less loss) while TransPro degrades them too. A matched-blur control (Table~\ref{tab:blur})
shows this is not lost resolution: the synthetic image carries as much high-frequency energy as real OCTA, but
in the wrong places.

\noindent\textbf{Contributions.}
\begin{itemize}
  \item We turn \emph{downstream-task fidelity} into a reusable, training-free probe for OCT-to-OCTA synthesis, a
  frozen real-OCTA segmenter that reads whether synthetic OCTA supports the vascular measurements OCTA is acquired
  for, stratified by vascular structure (large vessel, FAZ, capillary), and release it to score any synthesizer.
  \item We establish, across two state-of-the-art synthesizers (XOCT and TransPro), that synthesis collapses the
  fine capillary network (XOCT $0.798 \to 0.635$ Dice, five times the large-vessel loss; TransPro harder on
  every target), a failure invisible to reconstruction PSNR.
  \item Two controls localize the failure: a matched-blur control shows the missing capillaries are fabricated,
  not blurred, and a train-on-synthetic control shows the vascular information is absent, not merely
  domain-shifted.
\end{itemize}

\section{Related Work}
\label{sec:related}

\textbf{OCT-to-OCTA synthesis.} Cross-modal translation from OCT to OCTA has progressed from the first
deep-learning flow-map generators~\cite{lee2019flowmaps} and 2D conditional GANs to 3D volumetric models with
layer-aware supervision: TransPro~\cite{li2024transpro} adds a frozen
vessel-segmentation guidance network and projection consistency; XOCT~\cite{khosravi2025xoct} uses a 3D
multi-scale generator with cross-dimensional supervision and reports state-of-the-art OCTA-500 reconstruction;
MuTri~\cite{chen2025mutri} translates in a discrete latent space; diffusion and flow-matching variants continue
to appear~\cite{badhon2026bbdm}. Where these methods use vessel segmentation, they use it as training-time
\emph{guidance} to improve the synthesizer and report the outcome as pixel fidelity (MAE, PSNR, SSIM). The most
vascular-aware evaluations stop at on-image proxies computed from the synthetic angiogram itself, vessel-density
error~\cite{li2024transpro}, diabetic-retinopathy biomarker trends~\cite{badhon2026bbdm}, and density, caliber,
and tortuosity indices~\cite{badhon2024quant,chen2025curvilinear}, never a measurement made by a model trained
on real OCTA. None turns segmentation into a held-out \emph{evaluation} that asks whether the synthetic image
supports the measurement, stratifies by vascular structure, and isolates why it fails. We invert segmentation
from a signal that improves synthesis into a fixed probe that measures the utility gap, and apply it to two
state-of-the-art synthesizers of different families and years (XOCT and TransPro), so a gap is not attributable
to one weak model. Whether synthetic OCTA supports outer-retinal neovascular-lesion detection is, to our
knowledge, untested; we give a first qualitative, matched-vendor look and leave a powered study to future work
(Sec.~\ref{sec:discussion}).

\textbf{Task-based evaluation of synthesis.} That downstream task performance, not pixel fidelity, is the
discriminating measure of a medical image synthesizer is established outside our modality: segmentation-based
evaluation separates synthesizers that PSNR and SSIM rank as equivalent~\cite{metricsmatter2025}, task-based
protocols anchor cross-modal benchmarks such as ultrasound-to-MR~\cite{usmr2026} and are argued for across the
generative medical-imaging literature~\cite{genaimedsurvey2025}, and distribution-matching losses hallucinate
plausible but incorrect structure~\cite{cohen2018hallucinate}. In retina the concern surfaces in the reverse
direction: synthetic OCTA generated to \emph{train} vessel segmenters carries a sim-to-real
penalty~\cite{kreitner2024synth}. We apply this protocol to two published OCT-to-OCTA synthesizers, stratify the
gap by vascular structure, and add controls that separate fabrication from resolution and from segmenter domain
shift.

\textbf{OCTA vessel segmentation.} Supervised networks reach Dice above $0.85$ for vessel and FAZ
segmentation on real OCTA~\cite{ma2021rose,li2020octa500}; we use a frozen segmenter of this quality as the
fixed measuring instrument, not as a contribution.

\section{The Downstream-Task-Fidelity Protocol}
\label{sec:method}

Our contribution is a measurement protocol, not a model. It asks whether the clinical measurement OCTA is
acquired for survives synthesis, and reads the answer out of two frozen networks, so it belongs to the tradition
of task-based evaluation protocols rather than of new synthesizers. Nothing is trained inside the protocol.

\textbf{Ingredients.} A segmenter $S$ trained on \emph{real} OCTA en-face projections with pixel labels; an OCTA
synthesizer $G$ mapping a structural OCT volume $o$ to a synthetic OCTA volume $\hat v = G(o)$; and, for a
held-out eye, its real OCTA volume $v$ with label $y$. $S$ and $G$ are both frozen.

\textbf{Projection operator $P$.} $S$ reads en-face bands, so both volumes are reduced by the same operator $P$.
For band $b$ (inner-retinal, full, or outer-retinal), $P_b$ is the maximum-intensity projection over the depth
range of $b$ at each en-face location $(i,j)$, followed by the fixed orientation and percentile normalization of
the acquired projections:
\begin{equation}
P_b(u)(i,j) \;=\; \mathrm{norm}\!\Big( \max_{k \,\in\, b(i,j)} u(i,j,k) \Big),
\label{eq:proj}
\end{equation}
where $b(i,j)$ is the depth interval of band $b$ delimited by the eye's layer boundaries. Applying the
\emph{identical} $P$ to $v$ and $\hat v$ is what makes a gap test real-vs-synthetic and not a difference in
projection.

\textbf{Utility gap.} For target $t$ (large vessel, FAZ, capillary) read from band $b(t)$, the utility gap is
the drop in the segmentation's Dice overlap ($0$ = no overlap, $1$ = identical) when the input is synthetic
rather than real, under the same $S$ and $P$:
\begin{equation}
\Delta_t \;=\; \mathrm{Dice}\big(S(P_{b(t)}(v)),\, y_t\big) - \mathrm{Dice}\big(S(P_{b(t)}(\hat v)),\, y_t\big).
\label{eq:gap}
\end{equation}
$\Delta_t > 0$ means the synthetic image cannot support the measurement the real image supports. Reporting
$\Delta_t$ \emph{per target} is what localizes the failure: a benign global gap can hide a capillary collapse
(Sec.~\ref{sec:results}). The real arm $S(P_{b(t)}(v))$ is the matched baseline; for large vessels it reaches
Dice $0.862$, close to the $0.873$ on the device's provided projections, confirming $P$ is faithful. For the
finer FAZ and capillary targets the same $P$ lowers the real baseline further (FAZ $0.866 \to 0.764$, capillary
$0.833 \to 0.798$); because this applies identically to $v$ and $\hat v$, $\Delta_t$ stays a controlled
comparison.

\textbf{Two controls make the gap causal.} A raw $\Delta_t$ does not say \emph{why} synthesis fails; two fixed
operations separate the causes. \emph{(i) Matched blur}: let $\sigma^\star$ be the Gaussian width whose radial
power spectrum most closely matches that of the synthetic projection; if $\Delta_t$ were merely lost resolution,
real blurred by $\sigma^\star$ would segment as poorly as synthetic. \emph{(ii) Train on synthetic}: retrain
only $S$'s decoder on synthetic OCTA and re-measure; if $\Delta_t$ were a real-to-synthetic domain shift the gap
would close, if the vascular information is absent it does not.

\section{Experimental Setup}
\label{sec:setup}

\textbf{Data.} We use OCTA-500 (6\,mm, $400\times400$ en-face)~\cite{li2020octa500}, 300 eyes with paired
OCT/OCTA volumes, layer segmentations, and pixel vessel/FAZ labels. An en-face projection is a 2D map viewed
from the front, formed by collapsing the 3D flow signal through depth over a layer band. All downstream numbers
are on a 50-eye test split: the intersection of the synthesizer's 70-eye held-out test set with the segmenter's
held-out set (the 20 test eyes that fall in the segmenter's training are excluded). The three segmentation
targets are large vessel, FAZ, and capillary (the fine inner-retinal capillary meshwork whose dropout marks
diabetic ischemia); the three input bands are the inner-retinal (ILM--OPL), full-retinal, and outer-retinal
(OPL--BM, the outer retina below the deep vascular plexus and avascular in health) en-face projections, where
ILM, OPL, and BM are the inner limiting membrane,
outer plexiform layer, and Bruch's membrane retinal boundaries. For a cross-vendor test of neovascular pathology
(Sec.~\ref{sec:discussion}), we additionally use SpectralisMNV, a private dataset of $478$ paired OCT/OCTA
volumes on a Heidelberg Spectralis scanner ($3$--$4$\,mm macular fields, a different vendor and field from
OCTA-500's $6$\,mm), with per-eye diagnoses graded by a retina specialist.

\textbf{Synthesizers.} We use the released XOCT weights~\cite{khosravi2025xoct}, reproducing its reported
OCTA-500 6\,mm reconstruction to within $0.1$~dB PSNR and $0.003$ SSIM, and we train
TransPro~\cite{li2024transpro} from scratch on the same split to XOCT-comparable reconstruction
(Table~\ref{tab:repro}), so both are legitimately trained state-of-the-art models. MuTri~\cite{chen2025mutri}
releases code but not trained weights, so we could not reproduce its results and omit it.

\textbf{Segmenter.} The fixed probe is a frozen OCTCube encoder~\cite{octcube2024}, a Vision Transformer
pretrained by masked autoencoding on $26{,}685$ structural-OCT volumes, with a trained
feature-pyramid decoder that segments vessels and FAZ from real OCTA at Dice $0.873$/$0.866$/$0.833$ (large
vessel/FAZ/capillary), a level competitive with supervised baselines~\cite{ma2021rose,li2020octa500}. We
reproduce its large-vessel test Dice ($0.873$) in our setup before using it as a probe. Encoder, decoder, and
training configuration are released to make the probe reproducible.

\begin{table}[!t]
\centering
\caption{We reproduce XOCT's published OCTA-500 6\,mm reconstruction (70-eye test split) to within $0.1$~dB
PSNR and $0.003$ SSIM, and train a second synthesizer, TransPro, to XOCT-comparable reconstruction (within
$\sim$$1$--$1.6$~dB PSNR). Both run faithfully before we probe them, so a downstream gap reflects fidelity, not
undertraining. proj = en-face projection.}
\label{tab:repro}
\begin{tabular}{lcccc}
\toprule
 & 3D PSNR & 3D SSIM & proj-full PSNR & proj-full SSIM \\
\midrule
XOCT (reported) & 31.26 & 0.905 & 21.46 & 0.568 \\
XOCT (ours)     & 31.17 & 0.903 & 21.53 & 0.567 \\
TransPro (ours) & 29.54 & 0.880 & 20.78 & 0.532 \\
\bottomrule
\end{tabular}
\end{table}

\section{Synthesis Collapses the Fine Capillary Network}
\label{sec:results}

\textbf{The utility gap grows with structural fineness (Table~\ref{tab:strat}).} Running the frozen segmenter
on synthetic versus matched-real projections, large-vessel Dice barely moves ($0.862 \to 0.831$, a $4\%$
relative drop), FAZ degrades more ($0.764 \to 0.678$, $11\%$), and capillary segmentation collapses
($0.798 \to 0.635$, a $0.163$ absolute drop, five times the large-vessel loss). Every gap is significant
(paired Wilcoxon signed-rank over the 50 eyes: $p<10^{-3}$ for all three targets; $95\%$ bootstrap CIs on the
gap $[0.027, 0.036]$/$[0.043, 0.138]$/$[0.156, 0.170]$ for large vessel/FAZ/capillary); these intervals reflect
test-eye sampling at a fixed segmenter seed. At XOCT's matched full-projection PSNR of $21.5$~dB, a pixel-fidelity metric ranks the synthetic output as
strong while its downstream capillary utility differs from real by a factor of five. TransPro shows the same monotone
collapse with larger drops on every target (large vessel $0.101$, capillary $0.206$), so the utility gap is a
property of OCT-to-OCTA synthesis, not of one model. In
clinical units, the FAZ area read from the segmenter's mask corroborates the pattern: XOCT holds it within
measurement repeatability ($0.333$ vs $0.329$~mm$^2$ on real) while TransPro inflates it $\sim$$30\%$
($0.426$~mm$^2$), so the coarse FAZ survives
and the loss is specific to the fine capillary network.

\begin{table}[!t]
\centering
\caption{Downstream segmentation Dice on synthetic vs matched-recipe real OCTA (50 held-out eyes; Dice drop in
parentheses). Both synthesizers collapse monotonically with structural fineness at matched reconstruction PSNR:
XOCT preserves large vessels (drop $0.031$) while destroying capillaries ($5\times$), whereas TransPro degrades
even large vessels (drop $0.101$) and is more uniformly bad. Higher is better; the matched-real large-vessel
score ($0.862$) is close to the device's provided projections ($0.873$). All gaps are significant (paired
Wilcoxon signed-rank, $p<10^{-3}$); the XOCT capillary-gap $95\%$ CI is $[0.156, 0.170]$.}
\label{tab:strat}
\begin{tabular}{lccc}
\toprule
Segmentation target & Real (matched) & XOCT synth (drop) & TransPro synth (drop) \\
\midrule
Large vessel & 0.862 & 0.831 (0.031) & 0.761 (0.101) \\
Foveal avascular zone & 0.764 & 0.678 (0.086) & 0.609 (0.155) \\
Capillary & 0.798 & \textbf{0.635} (\textbf{0.163}) & \textbf{0.591} (\textbf{0.206}) \\
\bottomrule
\end{tabular}
\end{table}

\textbf{The collapse is structural, not lost resolution (Table~\ref{tab:blur}).} The obvious objection is that
synthesis simply produces a blurrier image. It does not, for either synthesizer. Both synthetic projections'
radial power spectra match real OCTA at only $\sigma=0.4$~px (essentially no blur), and real OCTA blurred to
that resolution still segments at $0.798$. Yet XOCT segments at $0.635$ and TransPro at $0.591$, a $20$--$26\%$
drop at matched resolution, and real OCTA reaches those levels only when blurred to $\sigma\!\approx\!1.0$--$1.2$
($2.5$--$3\times$ the synthetic's actual resolution). Both synthesizers are as sharp as real yet segment like
heavily blurred real: the high-frequency energy is present but misplaced. The fine vasculature is fabricated
(the retained high-frequency texture does not correspond to true capillaries), not smoothed away.

\textbf{The vascular signal is not recoverable by adaptation.} To separate genuine information loss from a
segmenter domain gap, we retrain the decoder directly on synthetic OCTA, paired with the real ground-truth
vessel labels, and test on synthetic, removing any real-to-synthetic distribution mismatch. The retrained decoder does not recover the capillary network: trained
and tested on synthetic, it reaches Dice $0.617$, below both the $0.635$ of the real-trained segmenter on
synthetic and the $0.798$ real baseline. A
segmenter that learns the synthetic distribution still cannot recover the capillary detail, so the gap reflects
information absent from the synthetic image, not a distribution mismatch the segmenter could adapt to. The
matched-blur control, which needs no retraining, is the primary evidence for structural fabrication. The collapse
is not an artifact of the foundation-model probe either: a from-scratch U-Net reproduces it (capillary
$0.796 \to 0.620$, drop $0.176$ vs the foundation probe's $0.163$; Suppl.\ Table~S1).

\begin{table}[!t]
\centering
\caption{Matched-blur control (capillary target). Both synthesizers are as sharp as real OCTA: their radial
power spectrum matches real OCTA blurred by only $\sigma=0.4$~px. Yet at that matched resolution the synthetic
loses 20--26\% of the capillary Dice, a gap resolution cannot explain; real OCTA must be blurred to
$\sigma\approx1.0$--$1.2$~px ($2.5$--$3\times$ the synthetic's own blur) before it segments as badly. The
collapse is structural fabrication, not lost resolution.}
\label{tab:blur}
\begin{tabular}{lccc}
\toprule
 & Sharpness ($\sigma$, px) & Capillary Dice & vs.\ real \\
\midrule
Real OCTA               & 0 (native)   & 0.798 & -- \\
XOCT synthetic          & 0.4 (= real) & 0.635 & $-20\%$ \\
TransPro synthetic      & 0.4 (= real) & 0.591 & $-26\%$ \\
\midrule
Real blurred to XOCT's Dice     & 1.0 & 0.635 & ($2.5\times$ more blur) \\
Real blurred to TransPro's Dice & 1.2 & 0.591 & ($3.0\times$ more blur) \\
\bottomrule
\end{tabular}
\end{table}

\section{Discussion and Future Work}
\label{sec:discussion}

\begin{figure}[!t]
\centering
\includegraphics[width=0.44\textwidth]{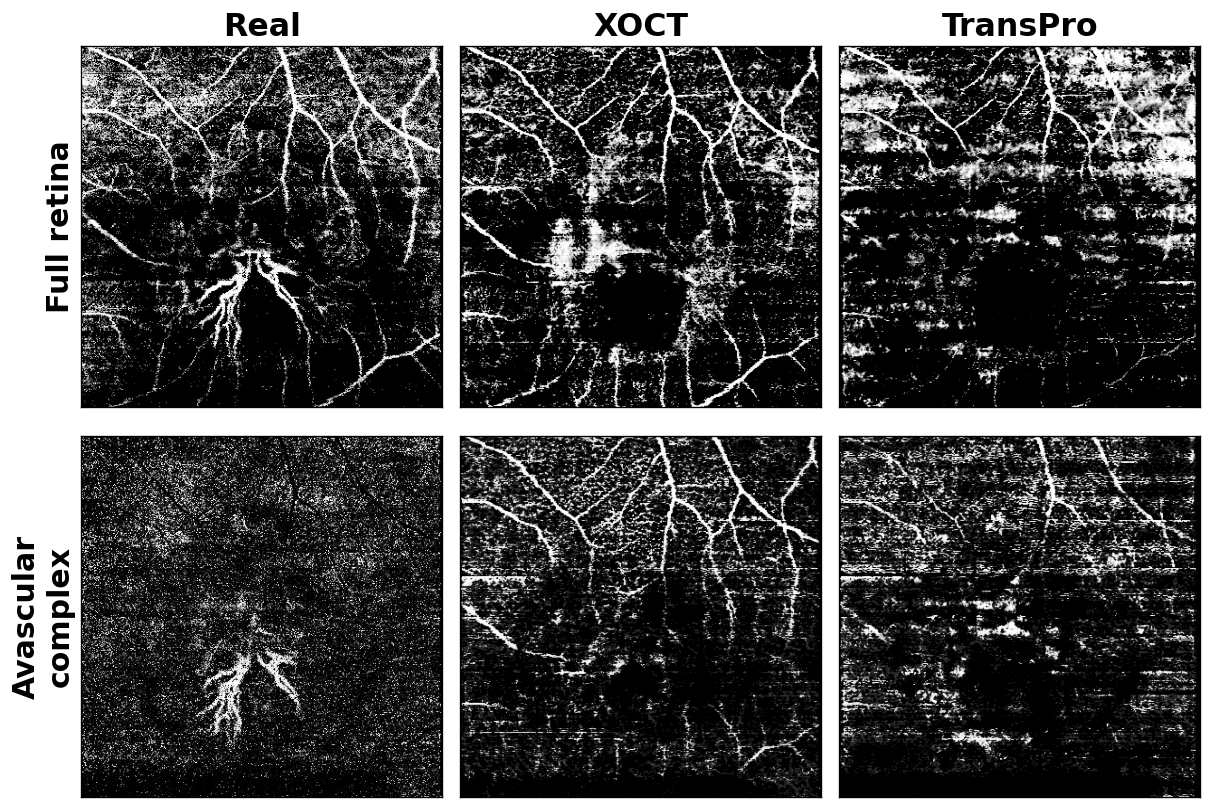}
\caption{\textbf{Neither synthesizer reproduces the neovascular lesion (\mbox{SpectralisMNV}).} Columns: real
OCTA, XOCT retrained on SpectralisMNV, TransPro retrained on the same data. Top row: full-retina en-face; bottom
row: the outer-retinal avascular-complex slab (OPL--BM), normally vessel-free
and thus the slab read for macular neovascularization (MNV). On the full retina both reproduce the
arcade vessels but leave a dark macular void; in the avascular complex the real carries a discrete branching
neovascular network (sea-fan / medusa morphology) that both synthesizers miss, rendering only the projected
superficial vessels. One representative eye of $n{=}3$ graded; illustrative, not a detection
metric.}
\label{fig:mnv}
\end{figure}

The capillary result is on a single OCTA engine (Optovue split-spectrum amplitude-decorrelation
angiography, SSADA) at $6$\,mm, and Dice is a surrogate for a clinical reading. Synthetic OCTA is not a
substitute for acquisition when the clinical read depends on capillary-level detail.

\textbf{The neovascular lesion of wet AMD.} The quantitative result above concerns the healthy capillary
network, a diabetic-retinopathy signal; the defining lesion of neovascular AMD is macular neovascularization
(MNV), pathological flow in the normally avascular outer-retinal slab (OPL--BM). The treatment
trigger is exudative fluid on structural OCT; OCTA detects, types, and monitors the network. OCTA-500 is poorly
suited to test MNV (few neovascular eyes, only the broad OPL--BM band), so we retrained both synthesizers on
SpectralisMNV (Heidelberg Spectralis, $478$ paired volumes at $3$--$4$\,mm) and held out eyes graded as showing
definite MNV. On the
three eyes with a confidently graded, discrete MNV network (one shown in Fig.~\ref{fig:mnv}; two typical MNV, one
pachychoroid neovasculopathy, a type-1 (sub-RPE) MNV complicating central serous chorioretinopathy), the real
outer slab carries a discrete
branching neovascular network that neither synthesizer reproduces: synthesis renders the population-typical outer
retina, not the eye-specific lesion. Both models still reproduce the arcade vessels (Fig.~\ref{fig:mnv}), so the
failure is specific to the eye's lesion, not a global reconstruction deficit from undertraining. The fair test is
localization, not exact morphology: the same anatomy-based extractor (Suppl.\ Alg.~S1) on either synthetic volume
recovers no lesion co-localizing with the real one (footprint IoU $0.00$--$0.18$). This qualitative case study is
scoped as such: a different vendor and field from the OCTA-500 arm, spectral-domain OCT that under-renders the
sub-RPE space where type-1 MNV sits, and reference masks that are clinician-accepted extractor outputs, not
independent tracings. A powered cohort scored against an independent tracing is the study our protocol enables.

\section{Conclusion}
\label{sec:conclusion}

Two state-of-the-art synthesizers both collapse the fine capillary network, a structural failure invisible to
reconstruction metrics. Reconstruction fidelity is not clinical utility: the vasculature OCTA is acquired to
measure should be the yardstick for the models that synthesize it.

\bibliographystyle{splncs04}
\bibliography{references}

\end{document}